\documentclass[sigconf]{acmart}

\usepackage{graphicx}
\usepackage{subcaption}
\usepackage{booktabs} 
\usepackage{hyperref}
\usepackage[capitalize,noabbrev]{cleveref}
\usepackage{url}            
\usepackage{nicefrac}       
\usepackage{multirow}
\usepackage{makecell}
\usepackage{enumitem}
\usepackage{xspace}
\usepackage{multicol}
\usepackage{array}
\usepackage{rotating}
\usepackage{soul}
\usepackage{colortbl}

\AtBeginDocument{%
  }

\setcopyright{acmlicensed}
\copyrightyear{2018}
\acmYear{2018}
\acmDOI{XXXXXXX.XXXXXXX}
\acmConference[Conference acronym 'XX]{Make sure to enter the correct
  conference title from your rights confirmation email}{June 03--05,
  2018}{Woodstock, NY}
\acmISBN{978-1-4503-XXXX-X/2018/06}

\newcommand{\circled}[1]{\textcircled{\scriptsize #1}}
\newcommand{\val}[3]{%
  \def\tempstd{#2}%
  \ifthenelse{\equal{\tempstd}{}}%
    {
      \ifthenelse{\equal{#3}{b}}%
        {\textbf{#1}}%
        {\ifthenelse{\equal{#3}{u}}%
          {\underline{#1}}%
          {#1}%
        }%
    }%
    {
      \ifthenelse{\equal{#3}{b}}%
        {\textbf{#1{\scriptsize $\pm$#2}}}%
        {\ifthenelse{\equal{#3}{u}}%
          {\underline{#1{\scriptsize $\pm$#2}}}%
          {#1{\scriptsize $\pm$#2}}%
        }%
    }%
}

\newif\ifcomm

\commtrue

\begin{document}

\title{Detecting Diffusion-Generated Time Series Under Generator Shift}

\author{Zhi Wen Soi}
\affiliation{%
  \institution{TU Dortmund University\\University of Neuch\^{a}tel}
  \city{Dortmund}
  \country{Germany}
}
\email{zhi.soi@tu-dortmund.de}

\author{Aditya Shankar}
\affiliation{%
  \institution{Delft University of Technology}
  \city{Delft}
  \country{Netherlands}
}
\email{a.shankar@tudelft.nl}

\author{Gert Lek}
\affiliation{%
  \institution{University of Neuch\^{a}tel}
  \city{Neuch\^{a}tel}
  \country{Switzerland}
}
\email{gert.lek@unine.ch}

\author{Abele M\u{a}lan}
\affiliation{%
 \institution{University of Neuch\^{a}tel}
 \city{Neuch\^{a}tel}
 \country{Switzerland}
}
\email{abele.malan@unine.ch}

\author{Daniel Neider}
\affiliation{%
  \institution{TU Dortmund University}
  \city{Dortmund}
  \country{Germany}
}
\email{daniel.neider@tu-dortmund.de}

\author{Jian-Jia Chen}
\affiliation{%
  \institution{RWTH Aachen University\\TU Dortmund University}
  \city{Aachen}
  \country{Germany}
}
\email{jian-jia.chen@rwth-aachen.de}

\author{Lydia Chen}
\affiliation{%
  \institution{University of Neuch\^{a}tel\\Delft University of Technology}
  \city{Neuch\^{a}tel}
  \country{Switzerland}
}
\email{yiyu.chen@unine.ch}

\renewcommand{\shortauthors}{Soi et al.}

\begin{abstract}
The boundary between real and diffusion-generated time series is becoming increasingly difficult to draw, yet detection in this domain remains underexplored, especially when the generator is unknown. We compare white-box detection, which requires access to the generator, against black-box detection, which operates on the raw signal alone. The white-box approach, a reconstruction-based detector adapted from the image domain, works well in in-distribution but breaks down under generator shift: reconstruction-based detection in images succeeds because large generic generators provide a near-universal reconstruction prior, and no analogous generator exists for time series. In contrast, a simple off-the-shelf classifier used as a black-box detector performs remarkably well, achieving an average F1 of 79.2, a \textbf{22.1\%} relative improvement over the white-box approach, and a TPR@1\%FPR of 57.2. Diffusion-generated time series detection is therefore not a direct transfer of the image domain problem.
This work provides the first systematic exploration of white-box and black-box detection for diffusion-generated time series. We close by identifying several open and promising directions.
Our code is available at \url{https://github.com/soizhiwen/DiffTSDetection}.

\end{abstract}




\maketitle

\begin{figure*}[ht!]
\captionsetup[subfigure]{justification=centering}
\centering
    \begin{subfigure}[b]{\linewidth}
        \centering
        \includegraphics[width=0.6\textwidth,trim={0cm 11cm 0cm 0cm},clip]{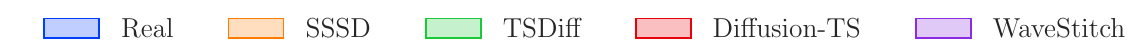}
    \end{subfigure}
    \begin{subfigure}[t]{0.325\linewidth}
        \centering
        \includegraphics[width=\textwidth]{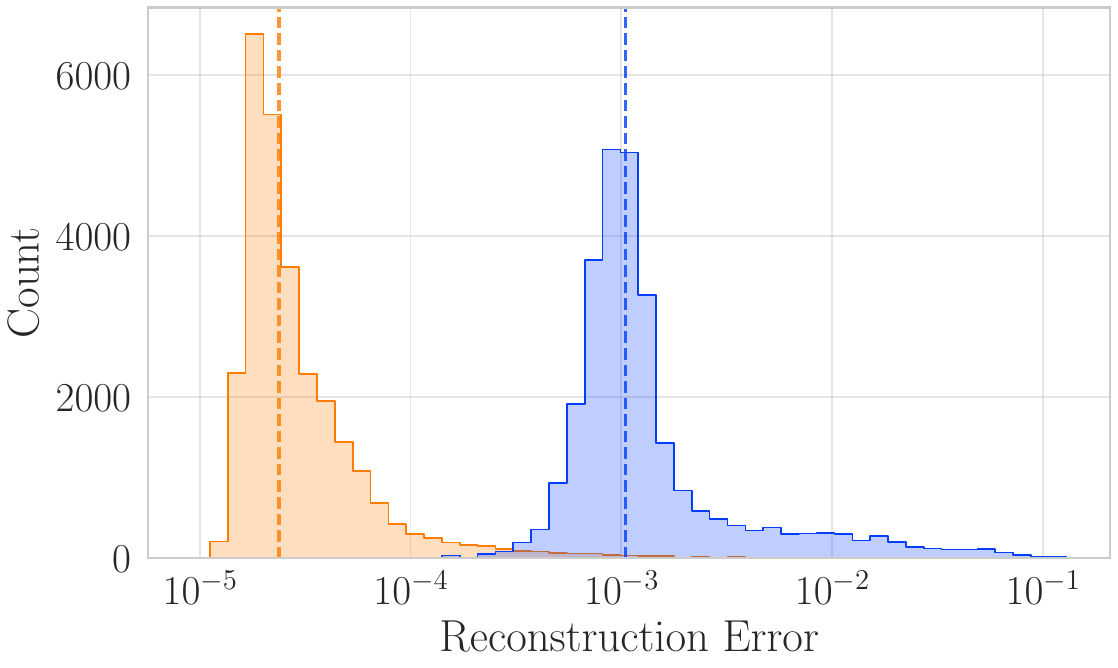}
        \caption{ETTm1 64-length (ID)}
        \label{subfig:pairs_dire_mse_ettm_64}
    \end{subfigure}
    \hfill
    \begin{subfigure}[t]{0.325\linewidth}
        \centering
        \includegraphics[width=\textwidth]{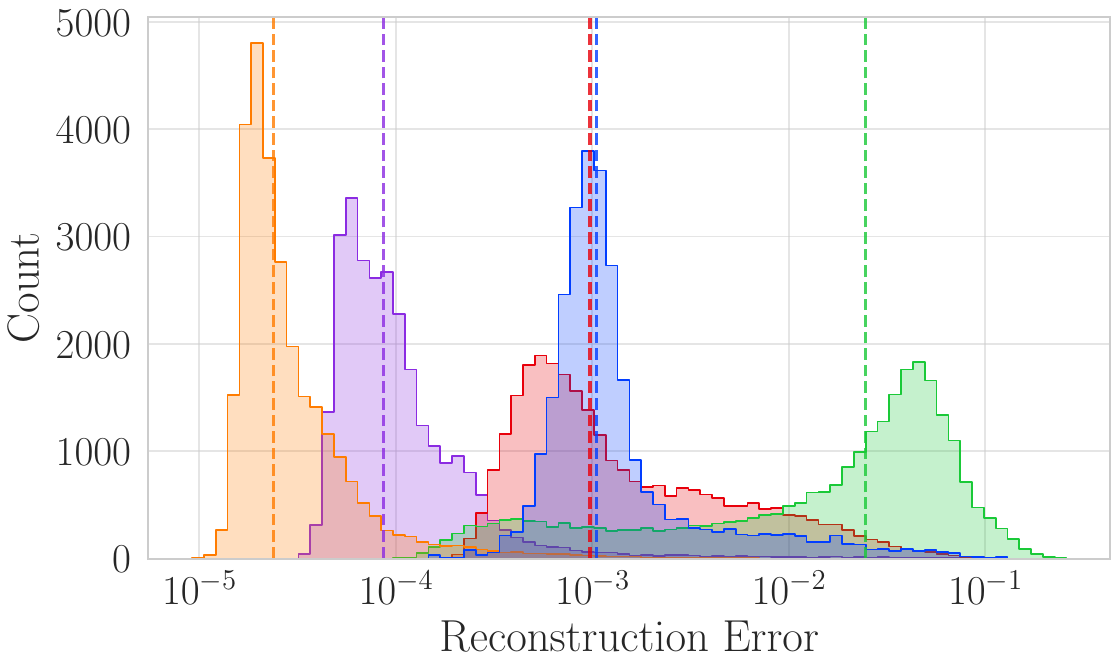}
        \caption{ETTm1 64-length (OOD)}
        \label{subfig:sssd_dire_mse_ettm_64}
    \end{subfigure}
    \hfill
    \begin{subfigure}[t]{0.325\linewidth}
        \centering
        \includegraphics[width=\textwidth]{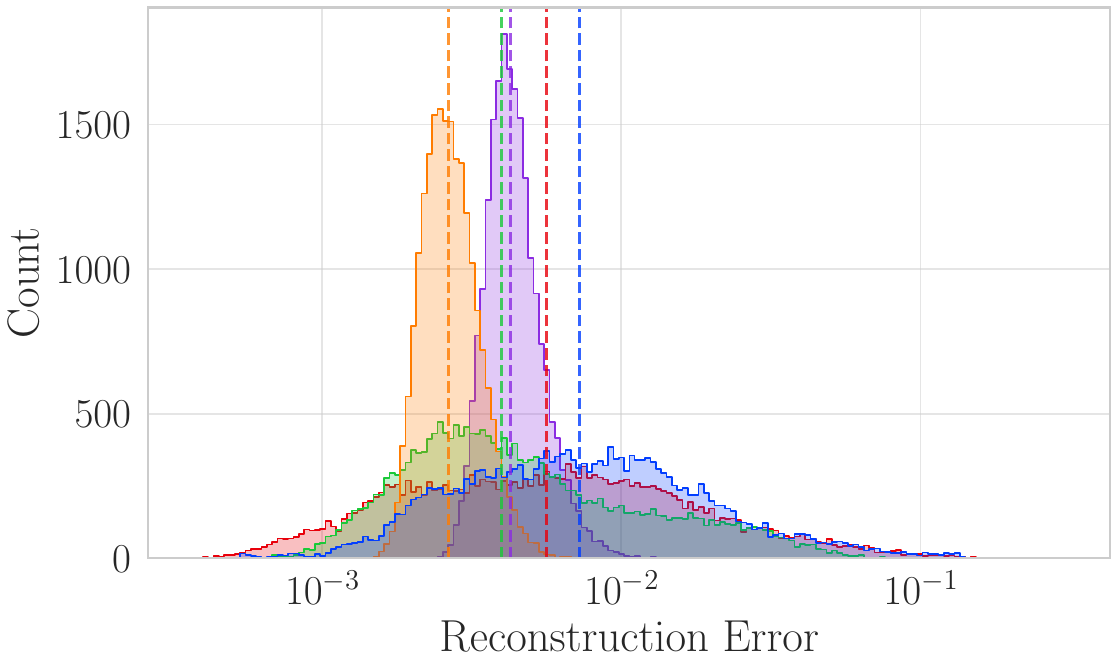}
        \caption{Weather 128-length (OOD)}
        \label{subfig:sssd_dire_mse_weather_128}
    \end{subfigure}
\caption{Reconstruction error distributions with SSSD used as the reconstruction model. (a) \textit{In-distribution case:} test samples come from SSSD. The reconstruction errors of real and synthetic samples are well separated, yielding a clear decision boundary. (b) and (c) \textit{Out-of-distribution case:} test samples include unseen generators (TSDiff, Diffusion-TS, and WaveStitch). While SSSD samples remain well or partially separated from real samples, the reconstruction errors of the unseen generators collapse onto the real samples, and the decision boundary disappears.}
\label{fig:dire_dist}
\Description{.}
\end{figure*}

\section{Introduction}

Time series data plays a crucial role in applications such as healthcare~\citep{DBLP:journals/jbi/MoridSKA20}, finance~\citep{DBLP:journals/access/LiCWLQY20}, and science~\citep{DBLP:journals/corr/abs-2406-16028}. However, obtaining large-scale real-world datasets remains challenging due to privacy constraints, limited accessibility, and the high cost of data collection. To alleviate these limitations, synthetic time series have emerged as a practical alternative for data sharing and model development~\citep{DBLP:journals/corr/abs-2403-04190, DBLP:conf/icml/DasKSZ24, DBLP:journals/corr/abs-2510-15821}. In particular, diffusion models~\citep{DBLP:conf/nips/HoJA20} have demonstrated strong performance in time series generation, often producing higher quality samples than traditional approaches such as generative adversarial networks and variational autoencoders~\citep{DBLP:conf/nips/KolloviehABZ0023, DBLP:conf/iclr/YuanQ24}. As a result, the boundary between real and diffusion-generated time series is becoming increasingly difficult to draw.

Distinguishing such synthetic time series from authentic ones thus becomes increasingly important. Reliable detection is essential for preventing misuse and maintaining trust in synthetic data applications~\citep{DBLP:conf/iccv/WangBZWHCL23, DBLP:conf/cvpr/RickerLF24}. However, unlike in the image domain, there is little prior work on detecting synthetic time series, especially when the generator is unknown. Existing work~\citep{DBLP:conf/aaai/HouZZW26} assumes the detector has direct access to the generator's internal distribution at inference time, which is often unavailable in realistic scenarios, and is moreover built for autoregressive generators rather than the diffusion-based ones we study. A detector is considered robust only when it can also identify synthetic samples from generators not seen during training~\citep{DBLP:conf/nips/Christian25}.

We study two strategies for detecting diffusion-generated time series. First, white-box detection requires access to a diffusion model trained on the same distribution as the generator of interest, mirroring the assumption behind image domain methods~\citep{DBLP:conf/iccv/WangBZWHCL23, DBLP:conf/cvpr/RickerLF24}. Black-box detection, in contrast, operates on the raw signal alone without access to the generator. We evaluate both strategies under two test conditions: in-distribution (ID), where test samples are generated by the same generator used during training, and out-of-distribution (OOD), where test samples are generated by unseen generators.

We find that the white-box approach, a reconstruction-based detector adapted from the image domain, works well in in-distribution. However, under generator shift, it breaks down. Reconstruction-based detection in the image domain succeeds in part because large generic generators such as Stable Diffusion~\citep{DBLP:conf/cvpr/RombachBLEO22} provide a reconstruction prior that approximately aligns with those of other image generators of interest. No analogous generic generator exists in the time series domain. Every available diffusion-based time series model is trained on a comparatively narrow distribution and learns model-specific reverse dynamics~\citep{DBLP:conf/nips/KolloviehABZ0023, DBLP:conf/iclr/YuanQ24}, so when the reconstruction model is applied to samples from a different generator, it corrects them toward its own learned distribution rather than exposing synthesis artifacts. The reconstruction error signal then collapses, and the reconstruction errors of synthetic and real samples become indistinguishable, as depicted in \cref{fig:dire_dist}.


Since the white-box approach relies on a generic reconstruction prior that does not exist in the time series domain, we revisit a simpler alternative that bypasses the reconstruction step entirely and learns directly from the raw signal. Therefore, unlike in the image domain, we find that an off-the-shelf classifier used as a black-box detector performs remarkably well. Across ten benchmark datasets, three sequence lengths, and four diffusion-based time series generators, a classic classifier trained on only samples generated from a single generator achieves an average OOD F1 of 79.2, a \textbf{22.1\%} relative improvement over the white-box approach, and a TPR@1\%FPR of 57.2.

Our results suggest that diffusion-generated time series detection is not a direct transfer from the image domain. Generator specificity introduces distinct challenges, and black-box detection is viable but remains poorly understood. To our knowledge, this work is the first systematic exploration of white-box and black-box detection for diffusion-generated time series, and we close by identifying several open directions. We summarize our contributions as follows:

\begin{enumerate}[leftmargin=2em]
    \item \textbf{First white-box detector for diffusion time series.} We adapt image domain reconstruction-based detection to time series and demonstrate its effectiveness in-distribution, achieving an average F1 of 91.9 across sequence lengths.
    \item \textbf{Failure of white-box detection under generator shift.} We find that the time series domain lacks a universal reconstruction prior: in the absence of generic generators, a reconstruction model maps samples from unseen generators toward its own learned distribution, making reconstruction errors for real and synthetic samples indistinguishable.
    \item \textbf{Black-box classifier is effective across generators.} An off-the-shelf time series classifier used as a black-box detector achieves an average F1 of 79.2, a \textbf{22.1\%} relative improvement over the white-box approach, and a TPR@1\%FPR of 57.2. These results demonstrate the viability of black-box detection and motivate further exploration of black-box time series detectors.
\end{enumerate}

\section{Related Work}

We review related work along three threads: time series generative models that produce the synthetic samples we aim to detect, classifier-based and reconstruction-based detection methods that motivate our black-box and white-box framing, and the limited existing work on detection in the time series domain.

\textbf{Time series generative models.} Time series generative models have advanced rapidly, with two dominant paradigms: diffusion-based and autoregressive generation. Diffusion-based generators iteratively denoise samples from a Gaussian prior to produce realistic time series. Recent examples include SSSD~\citep{DBLP:journals/tmlr/AlcarazS23}, TSDiff~\citep{DBLP:conf/nips/KolloviehABZ0023}, Diffusion-TS~\citep{DBLP:conf/iclr/YuanQ24}, and WaveStitch~\citep{DBLP:journals/pacmmod/ShankarCDH25}. Autoregressive generators, by contrast, produce time series step-by-step, and recent time series foundation models such as MOMENT~\citep{DBLP:conf/icml/GoswamiSCCLD24}, TimesFM~\citep{DBLP:conf/icml/DasKSZ24}, and Chronos~\citep{DBLP:journals/corr/abs-2510-15821} follow this paradigm with large transformer backbones. We focus on detecting diffusion-based generators in this work.

\textbf{Classifier-based detection.} Classifier-based methods, which are black-box in our context, train a detector directly on real and generated samples. CNNSpot~\citep{DBLP:conf/cvpr/WangW0OE20} shows that a detector trained on generated images can generalize to unseen generators with proper image augmentation. More recent approaches leverage large pre-trained vision models as feature extractors followed by a classifier~\citep{DBLP:conf/cvpr/OjhaLL23, DBLP:conf/iclr/YanLCHJ0X25, DBLP:conf/nips/Christian25}. However, these methods are designed around image-specific features and do not directly transfer to time series.

\textbf{Reconstruction-based detection.} Reconstruction-based methods, which are white-box in our case, utilize the generator itself and provide a robust signal for detection. Such methods require a diffusion model aligned with the generator that produced the test samples. DIRE~\citep{DBLP:conf/iccv/WangBZWHCL23} shows that diffusion-generated images reconstruct more accurately than real ones under a pre-trained diffusion model. AEROBLADE~\citep{DBLP:conf/cvpr/RickerLF24} uses autoencoder reconstruction errors from latent diffusion models, but most state-of-the-art (SOTA) time series diffusion models operate directly in data space rather than through a latent representation~\citep{DBLP:conf/nips/KolloviehABZ0023, DBLP:conf/iclr/YuanQ24}, making AEROBLADE inapplicable to time series.

\textbf{Detection in the time series domain.} In contrast, detection in the time series domain remains underexplored. UCE~\citep{DBLP:conf/aaai/HouZZW26} introduces a white-box framework tailored to autoregressive large model-generated time series, leveraging direct access to the generator's internal distribution to construct a theoretically grounded detection criterion. Beyond requiring generator access at detection time, which is often unrealistic in practice, UCE is built for autoregressive forecasting models rather than the broader diffusion-based generators we study here.
Since synthetic time series detection is naturally a binary classification problem, we instead study various off-the-shelf time series classifiers~\citep{DBLP:journals/datamine/FawazLFPSWWIMP20, DBLP:conf/dsaa/IsmailFawazDBWF23, DBLP:conf/incdm/FoumaniTS21} as our black-box detector that takes the raw signal as input.
\section{Diffusion-Generated Time Series Detection} \label{sec:method}

We present two strategies for detecting diffusion-generated time series: a white-box reconstruction-based detector adapted from the image domain in \cref{subsec:whitebox}, and a black-box classifier for the cross-generator setting in \cref{subsec:crossgenerator}. We show that the white-box approach works well in the in-distribution case but fails under generator shift, motivating the black-box alternative.

\subsection{White-Box Reconstruction-Based Detection} \label{subsec:whitebox}

We adapt DIRE~\citep{DBLP:conf/iccv/WangBZWHCL23}, originally proposed in the image domain, to time series as our white-box detector. The defining property of the white-box setting is access to a reference generator \(G^*\): a pre-trained diffusion model that produces synthetic samples \(x^* \in \mathbb{R}^{L \times C}\) drawn from its learned distribution \(p_{G^*}\), where \(L\) is the sequence length and \(C\) is the number of channels. We repurpose this same diffusion model as the reconstruction model and ask, for some arbitrary input, whether \(G^*\) can reconstruct it.

Given an input time series \(x_0 \in \mathbb{R}^{L \times C}\), the reconstruction model first applies Denoising Diffusion Implicit Model (DDIM) inversion~\citep{DBLP:conf/iclr/SongME21} to progressively add Gaussian noise and obtain a noisy state \(x_T\). It then denoises \(x_T\) back through the same diffusion model to produce a reconstructed sample \(\hat{x}_0 = \mathcal{R}_{G^*}(x_0)\), where \(\mathcal{R}_{G^*}\) denotes the full DDIM inversion and denoising process under \(G^*\). We then define the reconstruction error under \(G^*\) as:
\[
d_{G^*}(x_0) = \left(x_0 - \mathcal{R}_{G^*}(x_0)\right)^2,
\]
which is used as a detection cue. Then a downstream binary classifier maps \(d_{G^*}(x_0)\) to a real or synthetic label.

The intuition is straightforward. For a synthetic sample \(x^* \sim p_{G^*}\), it lies close to the reverse process trajectory the reconstruction model has learned, so \(\mathcal{R}_{G^*}(x^*)\) stays close to \(x^*\) and \(d_{G^*}(x^*)\) is small. Real samples are not aligned with this trajectory, so their reconstruction errors are larger. The two distributions are therefore well separated in the in-distribution case, as shown in \cref{subfig:pairs_dire_mse_ettm_64}, and a simple downstream classifier can recover the decision boundary.

\subsection{Cross-Generator Detection} \label{subsec:crossgenerator}

\begin{figure}
    \centering
    \includegraphics[width=\linewidth]{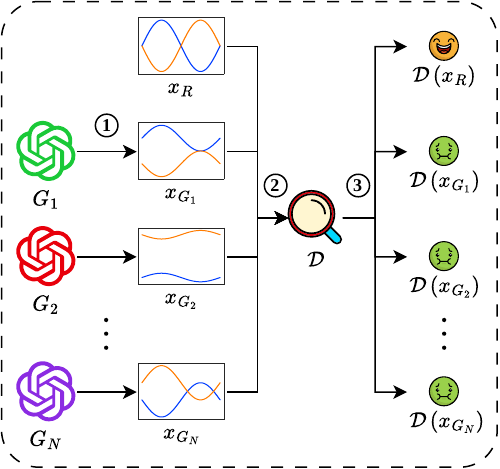}
    \caption{Cross-generator detection. \circled{1} Each generator \(G_i \in \mathcal{G} = \{G_1, G_2, \dots, G_N\}\) generates a synthetic sample \(x_{G_i}\). \circled{2} The detector $\mathcal{D}$ receives a real sample \(x_R\) along with synthetic samples and \circled{3} outputs a binary prediction (real or synthetic) for each.}
    \label{fig:overview}
    \Description{.}
\end{figure}

In a realistic detection scenario, we encounter arbitrary time series generated by unknown models. We refer to this as \textit{cross-generator detection}, an OOD setting where test samples may come from generators not seen during training.

\textbf{Setup.} \cref{fig:overview} illustrates such a scenario. Beyond the reference generator \(G^*\) introduced in \cref{subsec:whitebox}, we consider real time series \(x_R \in \mathbb{R}^{L \times C}\) drawn from the real distribution \(p_R\), and a set of unseen generators \(\mathcal{G} = \{G_1, G_2, \dots, G_N\}\), where each \(G_i \in \mathcal{G}\) produces a synthetic sample \(x_{G_i} \in \mathbb{R}^{L \times C}\) from its own distribution \(p_{G_i}\). In black-box detection, \(G^*\) also provides the labeled synthetic samples used to train the classifier. Crucially, the detector has no exposure to any \(G_i \in \mathcal{G}\) during setup. The detector \(\mathcal{D}\) takes an input \(x \in \{x_R, x_{G_1}, \dots, x_{G_N}\}\) and outputs a binary prediction \(\hat{y} \in \{0, 1\}\), with \(\hat{y} = 0\) for real and \(\hat{y} = 1\) for synthetic.

\begin{figure}
    \centering
    \includegraphics[width=\linewidth]{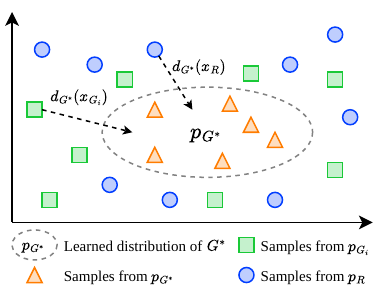}
    \caption{The reconstruction model has learned \(p_{G^*}\) specific to the reference generator \(G^*\), so the reconstruction error effectively measures distance to \(p_{G^*}\). Samples from an unseen generator \(G_i\) are off-distribution for \(G^*\) in model-specific ways, and real samples are likewise off-distribution. The two distances \(d_{G^*}(x_{G_i})\) and \(d_{G^*}(x_R)\) are of comparable magnitude, so their reconstruction errors overlap and the detector cannot separate synthetic from real.}
    \label{fig:manifold}
    \Description{.}
\end{figure}

\textbf{Why reconstruction-based detection fails in time series.} The reconstruction model is trained to invert and reconstruct samples from \(p_{G^*}\), so it has learned the structure of \(p_{G^*}\) together with the reverse process dynamics that map nearby points back to it. For any input \(x\), \(d_{G^*}(x)\) measures the distance from \(x\) to \(p_{G^*}\).

For a sample \(x_{G_i} \sim p_{G_i}\) from an unseen generator \(G_i \in \mathcal{G}\), \(G_i\) approximates the same underlying real distribution as \(G^*\) but learns a different distribution \(p_{G_i} \neq p_{G^*}\). The DDIM inversion and reconstruction path \(x_{G_i} \rightarrow x_T \rightarrow \hat{x}_{G_i}\) only recovers \(x_{G_i}\) accurately when \(x_{G_i}\) is compatible with the reverse dynamics of \(G^*\). Since \(x_{G_i}\) lies outside \(p_{G^*}\) in model-specific ways, the reconstruction model corrects it toward \(p_{G^*}\) rather than reconstructing it. Real samples \(x_R\) likewise lie outside \(p_{G^*}\), with \(d_{G^*}(x_{R})\) of comparable magnitude to \(d_{G^*}(x_{G_i})\). \cref{fig:manifold} illustrates this, when \(d_{G^*}(x_{G_i})\) and \(d_{G^*}(x_{R})\) overlap, and the reconstruction error no longer separates synthetic from real samples. The failure becomes even more obvious when the underlying distributions \(p_R\), \(p_{G_i}\), and \(p_{G^*}\) themselves substantially overlap. In other words, the reconstruction errors of all three collapse onto each other, leaving no decision boundary at all, as highlighted in \cref{subfig:sssd_dire_mse_weather_128}. In short, the white-box detector measures distance to \(p_{G^*}\), not the synthetic or real distinction, and this distance is uninformative under generator shift.

\begin{figure}
    \centering
    \includegraphics[width=0.6\linewidth]{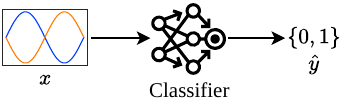}
    \caption{Black-box classifier-based detection. The classifier takes a raw time series \(x\) as input and predicts whether it is real or synthetic, with no generator access at inference time.}
    \label{fig:disjointcnn_pipeline}
    \Description{.}
\end{figure}

\textbf{Black-box classifier-based detection.} To avoid the generator specificity problem in time series, we revisit a simpler alternative: train a classifier directly on labeled real and diffusion-generated time series, bypassing the reconstruction step entirely. As illustrated in \cref{fig:disjointcnn_pipeline}, a raw time series signal \(x\) is fed into the classifier, which directly outputs a prediction \(\hat{y}\). The training minimizes the standard binary cross-entropy objective as defined as follows:
\[
\mathcal{L}(y,\hat{y}) = -\sum_{i=1}^B \left[ y_i \log(\hat{y}_i) + (1-y_i) \log(1-\hat{y}_i) \right],
\]
where $B$ denotes the mini-batch size, \(y\) is the actual label, and \(\hat{y}\) is the classifier's prediction. The classifier is trained only on labeled samples drawn from \(p_R\) and \(p_{G^*}\).

The black-box approach does not rely on any reconstruction prior and is hard to interpret; however, it is not affected by the generator-specifity problem that breaks white-box detection. The classifier learns synthesis artifacts directly from raw samples, and these artifacts can transfer across generators when different diffusion models leave overlapping fingerprints in their samples.

\textbf{Architecture choices.} There are many time series classifiers that can be used as black-box detectors. TimeCNN~\citep{DBLP:journals/jsee/Zhao17} and FCN~\citep{DBLP:conf/ijcnn/WangYO17} capture local temporal dependencies through convolutional filters, while MLP~\citep{DBLP:conf/ijcnn/WangYO17} serves as a simple baseline without explicit temporal structure. ResNet~\citep{DBLP:conf/ijcnn/WangYO17} improves capacity via residual connections, and Disjoint-CNN~\citep{DBLP:conf/incdm/FoumaniTS21} extends convolutional modeling to multivariate inputs by separating temporal and channel-wise convolutions. More recent approaches, such as LITE~\citep{DBLP:conf/dsaa/IsmailFawazDBWF23} and InceptionTime~\citep{DBLP:journals/datamine/FawazLFPSWWIMP20}, adopt multi-scale designs to capture patterns across different temporal resolutions. In this work, we focus primarily on Disjoint-CNN due to its effectiveness at identifying subtle artifacts in synthetic time series.

\section{Experiments}

In this section, we present the empirical results of our study and analyze the factors that lead to the performance gap between classifier-based and reconstruction-based detectors.

\subsection{Experimental Settings}


\begin{table}
\centering
\caption{Datasets used for evaluation, along with their number of channels and total length.}
\label{tab:datasets}
\begin{tabular}{lrr}
\toprule
Name & Channels & Length \\
\midrule
Electricity & 321 & 26 304 \\
Energy & 28 & 19 735 \\
ETTh1 & 7 & 17 420 \\
ETTm1 & 7 & 69 680 \\
Exchange Rate & 8 & 7 588 \\
fMRI & 50 & 10 000 \\
Illness & 7 & 966 \\
Stock & 6 & 3 685 \\
Traffic & 862 & 17 544 \\
Weather & 21 & 52 696 \\
\bottomrule
\end{tabular}
\end{table}

\textbf{Setup.} We evaluate the detectors on ten benchmark datasets, with details provided in \cref{tab:datasets}. Reported results are averaged across all datasets. We consider four SOTA diffusion-based time series generators: \textit{SSSD}~\citep{DBLP:journals/tmlr/AlcarazS23}, \textit{TSDiff}~\citep{DBLP:conf/nips/KolloviehABZ0023}, \textit{Diffusion-TS}~\citep{DBLP:conf/iclr/YuanQ24}, and \textit{WaveStitch}~\citep{DBLP:journals/pacmmod/ShankarCDH25} (more details in \cref{appx:experimental_settings}). For the white-box detector, we use DIRE~\citep{DBLP:conf/iccv/WangBZWHCL23}, as described in \cref{subsec:whitebox}. For the black-box detector, we consider three SOTA time series classifiers: InceptionTime~\citep{DBLP:journals/datamine/FawazLFPSWWIMP20}, LITE~\citep{DBLP:conf/dsaa/IsmailFawazDBWF23}, and Disjoint-CNN~\citep{DBLP:conf/incdm/FoumaniTS21}. Among these, Disjoint-CNN achieves the strongest average OOD performance (see \cref{tab:exp_sssd} in \cref{appx:classifiers}) and is therefore used throughout the main results.

\textbf{Training and testing settings.} SSSD serves as the in-distribution generator \(G^*\), since it is the earliest of the four considered models. This choice mirrors the realistic scenario in which a detector is built on an existing generator and later evaluated against newer or unseen ones. Concretely, DIRE uses SSSD's pre-trained diffusion model as the reconstruction model, while Disjoint-CNN is trained on real samples paired with SSSD-generated synthetic samples. Results are reported under two test conditions: (1) \textit{In-Distribution} (ID), where detectors are evaluated on the out-of-sample test split from SSSD; and (2) \textit{Out-of-Distribution} (OOD), where detectors are tested on samples from the three newer and unseen generators (TSDiff, Diffusion-TS, and WaveStitch).

\textbf{Evaluation metrics.} Detection performance is evaluated using \textit{F1-score}, \textit{Accuracy}, \textit{Average Precision} (AP), \textit{Area Under the Curve} (AUC), and \textit{True Positive Rate at a 1\% False Positive Rate} (TPR@1\%FPR). We provide
additional details on these metrics in \cref{appx:experimental_settings}.


\begin{table}
\caption{All metrics are higher the better, with TPR measured at a 1\% FPR. Values in \textcolor{gray}{gray} are ID, while values in black are OOD. The best average OOD scores are highlighted in \colorbox{green!15}{green}.}
\label{tab:main_exp}
\resizebox{\columnwidth}{!}{%
\begin{tabular}{llccccc}

\toprule

Generator & Method & F1 $\uparrow$ & Acc. $\uparrow$ & AP $\uparrow$ & AUC $\uparrow$ & TPR $\uparrow$ \\

\midrule
\multicolumn{7}{c}{\cellcolor{blue!15}{\textit{32-length}}} \\
\midrule

\multirow{2}{*}{SSSD} & DIRE & \textcolor{gray}{94.2} & \textcolor{gray}{95.4} & \textcolor{gray}{99.3} & \textcolor{gray}{99.2} & \textcolor{gray}{93.0} \\
& Disjoint-CNN & \textcolor{gray}{99.7} & \textcolor{gray}{99.7} & \textcolor{gray}{100.} & \textcolor{gray}{100.} & \textcolor{gray}{99.8} \\
\cmidrule{1-7}
\multirow{2}{*}{TSDiff} & DIRE & 62.8 & 47.1 & 52.0 & 48.9 & 0.00 \\
& Disjoint-CNN & 84.1 & 77.8 & 95.2 & 95.4 & 76.5 \\
\cmidrule{1-7}
\multirow{2}{*}{Diffusion-TS} & DIRE & 68.8 & 57.8 & 58.7 & 59.9 & 0.00 \\
& Disjoint-CNN & 71.7 & 59.2 & 61.9 & 57.8 & 19.6 \\
\cmidrule{1-7}
\multirow{2}{*}{WaveStitch} & DIRE & 76.4 & 69.9 & 77.5 & 79.1 & 10.0 \\
& Disjoint-CNN & 98.5 & 98.3 & 99.9 & 99.9 & 99.2 \\
\cmidrule{1-7}\morecmidrules\cmidrule{1-7}
\multirow{2}{*}{Avg. OOD} & DIRE & 69.3 & 58.2 & 62.7 & 62.6 & 3.33 \\
& Disjoint-CNN & \cellcolor{green!15}{84.8} & \cellcolor{green!15}{78.4} & \cellcolor{green!15}{85.7} & \cellcolor{green!15}{84.4} & \cellcolor{green!15}{65.1} \\

\midrule
\multicolumn{7}{c}{\cellcolor{blue!15}{\textit{64-length}}} \\
\midrule

\multirow{2}{*}{SSSD} & DIRE & \textcolor{gray}{93.4} & \textcolor{gray}{94.6} & \textcolor{gray}{98.6} & \textcolor{gray}{98.1} & \textcolor{gray}{81.7} \\
& Disjoint-CNN & \textcolor{gray}{99.9} & \textcolor{gray}{99.9} & \textcolor{gray}{99.9} & \textcolor{gray}{100.} & \textcolor{gray}{98.6} \\
\cmidrule{1-7}
\multirow{2}{*}{TSDiff} & DIRE & 66.6 & 55.0 & 59.7 & 60.8 & 0.00 \\
& Disjoint-CNN & 84.6 & 77.7 & 98.9 & 99.2 & 86.4 \\
\cmidrule{1-7}
\multirow{2}{*}{Diffusion-TS} & DIRE & 66.0 & 54.0 & 55.5 & 56.4 & 0.00 \\
& Disjoint-CNN & 71.5 & 59.3 & 67.6 & 65.5 & 15.7 \\
\cmidrule{1-7}
\multirow{2}{*}{WaveStitch} & DIRE & 67.1 & 55.6 & 63.5 & 63.9 & 0.00 \\
& Disjoint-CNN & 95.4 & 93.7 & 97.2 & 98.1 & 81.6 \\
\cmidrule{1-7}\morecmidrules\cmidrule{1-7}
\multirow{2}{*}{Avg. OOD} & DIRE & 66.6 & 54.9 & 59.6 & 60.4 & 0.00 \\
& Disjoint-CNN & \cellcolor{green!15}{83.8} & \cellcolor{green!15}{76.9} & \cellcolor{green!15}{87.9} & \cellcolor{green!15}{87.6} & \cellcolor{green!15}{61.2} \\

\midrule
\multicolumn{7}{c}{\cellcolor{blue!15}{\textit{128-length}}} \\
\midrule

\multirow{2}{*}{SSSD} & DIRE & \textcolor{gray}{88.0} & \textcolor{gray}{93.1} & \textcolor{gray}{95.2} & \textcolor{gray}{95.5} & \textcolor{gray}{89.9} \\
& Disjoint-CNN & \textcolor{gray}{94.7} & \textcolor{gray}{96.5} & \textcolor{gray}{100.} & \textcolor{gray}{100.} & \textcolor{gray}{100.} \\
\cmidrule{1-7}
\multirow{2}{*}{TSDiff} & DIRE & 58.6 & 45.9 & 50.0 & 46.6 & 0.00 \\
& Disjoint-CNN & 63.7 & 49.8 & 81.9 & 80.0 & 52.7 \\
\cmidrule{1-7}
\multirow{2}{*}{Diffusion-TS} & DIRE & 58.3 & 43.5 & 44.1 & 36.5 & 0.00 \\
& Disjoint-CNN & 62.8 & 46.9 & 46.7 & 36.2 & 10.2 \\
\cmidrule{1-7}
\multirow{2}{*}{WaveStitch} & DIRE & 59.6 & 50.4 & 55.3 & 56.4 & 2.01 \\
& Disjoint-CNN & 80.8 & 75.9 & 93.5 & 93.0 & 73.1 \\
\cmidrule{1-7}\morecmidrules\cmidrule{1-7}
\multirow{2}{*}{Avg. OOD} & DIRE & 58.8 & 46.6 & 49.8 & 46.5 & 0.67 \\
& Disjoint-CNN & \cellcolor{green!15}{69.1} & \cellcolor{green!15}{57.5} & \cellcolor{green!15}{74.1} & \cellcolor{green!15}{69.7} & \cellcolor{green!15}{45.3} \\

\bottomrule
\end{tabular}
}
\end{table}

\subsection{Results}

\begin{figure*}[ht!]
\captionsetup[subfigure]{justification=centering}
\centering
    \begin{subfigure}[b]{\linewidth}
        \centering
        \includegraphics[width=0.8\textwidth,trim={0cm 14cm 0cm 0cm},clip]{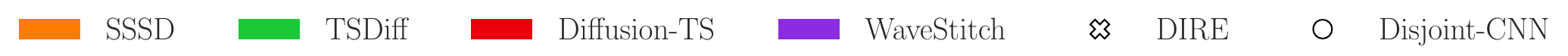}
    \end{subfigure}
    \begin{subfigure}[t]{0.325\linewidth}
        \centering
        \includegraphics[width=\textwidth]{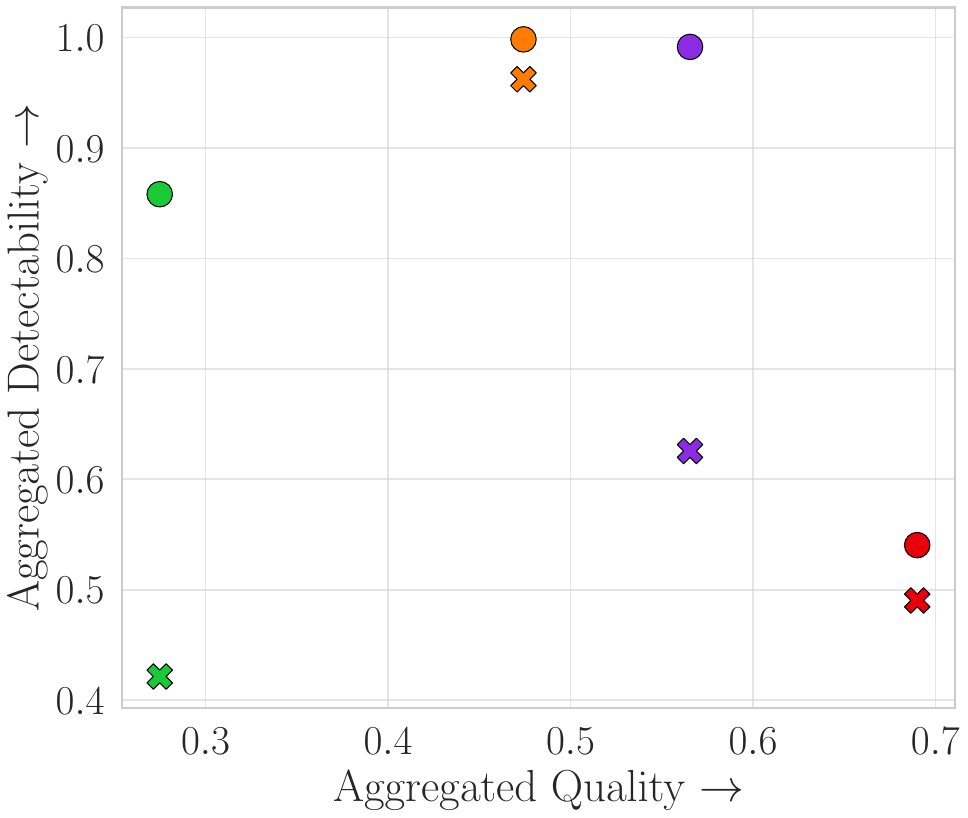}
        \caption{32-length}
        \label{subfig:quality_detectability_scatter_32}
    \end{subfigure}
    \hfill
    \begin{subfigure}[t]{0.325\linewidth}
        \centering
        \includegraphics[width=\textwidth]{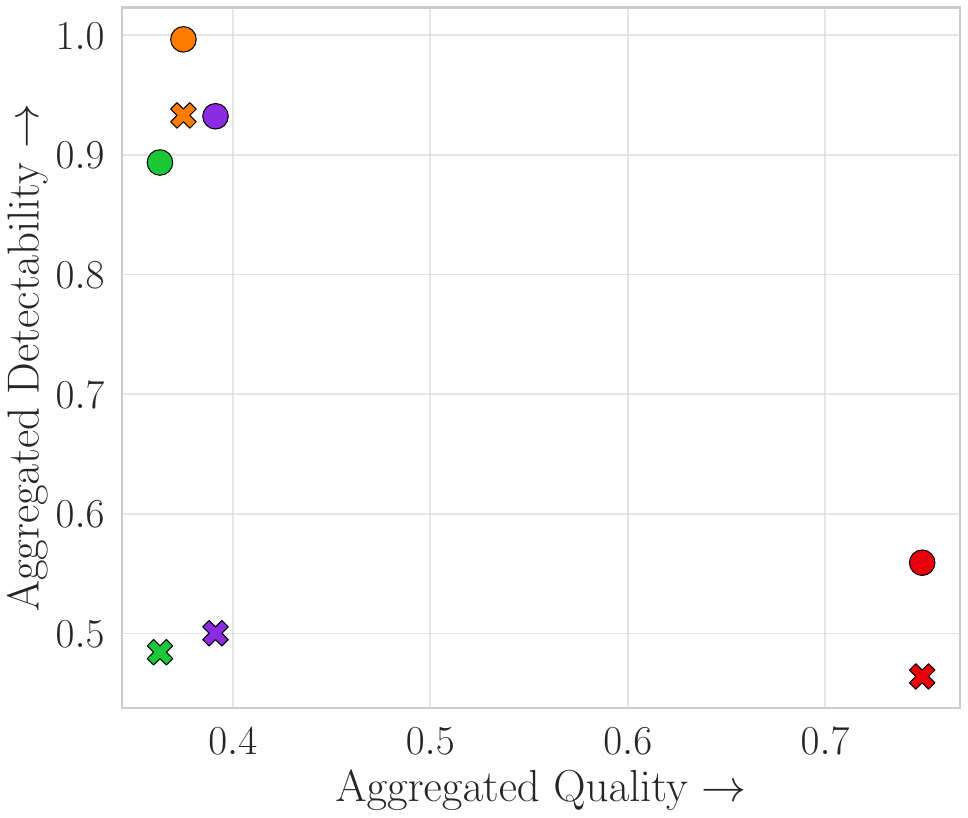}
        \caption{64-length}
        \label{subfig:quality_detectability_scatter_64}
    \end{subfigure}
    \hfill
    \begin{subfigure}[t]{0.325\linewidth}
        \centering
        \includegraphics[width=\textwidth]{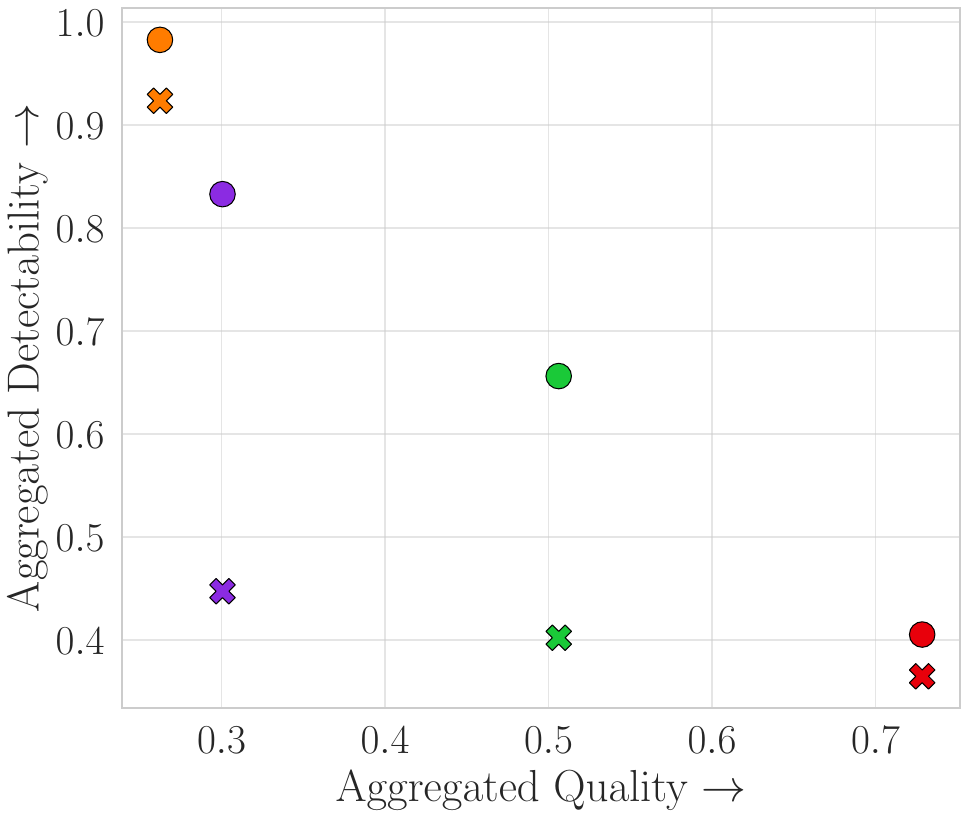}
        \caption{128-length}
        \label{subfig:quality_detectability_scatter_128}
    \end{subfigure}
\caption{Aggregated detectability versus aggregated quality for each generator across ten datasets at different sequence lengths. Aggregated detectability is the mean of all detectability metrics. Aggregated quality averages Context-FID~\citep{DBLP:conf/iclr/JehaBMKNFGJ22}, Correlational~\citep{DBLP:journals/corr/abs-2006-05421}, Discriminative~\citep{DBLP:conf/nips/YoonJS19}, and Predictive~\citep{DBLP:conf/nips/YoonJS19} scores, each normalized across the ten datasets and inverted so that higher values indicate higher quality. Higher synthetic data quality generally leads to lower detectability. Disjoint-CNN maintains consistently higher detectability than DIRE across all generators and sequence lengths.}
\label{fig:quality_detectability}
\Description{.}
\end{figure*}

\textbf{In-distribution.} In \cref{tab:main_exp}, Disjoint-CNN achieves near-perfect in-distribution performance on the out-of-sample test split from SSSD, with F1 above 94 across all three sequence lengths. DIRE also performs strongly, with an average ID F1 of 91.9, but falls notably short of this ceiling. Despite the reconstruction model being aligned with the generator that produced the test samples, DIRE cannot perfectly separate real from synthetic samples. As shown in \cref{subfig:pairs_dire_mse_ettm_64}, the reconstruction error distributions of real and SSSD-generated samples partially overlap, since DDIM inversion is an approximation process that introduces variability in reconstruction error for both real and synthetic inputs~\citep{DBLP:conf/iclr/SongME21, DBLP:conf/nips/SoiZhu25}. Disjoint-CNN operates directly on the raw time series, where synthesis artifacts remain more distinguishable, and is therefore not subject to this approximation error.

\textbf{Out-of-distribution.} The performance gap widens substantially when both detectors face unseen generators. Disjoint-CNN achieves an average OOD F1 of 79.2 and TPR@1\%FPR of 57.2 across the three sequence lengths, a 22.1\% relative improvement over DIRE's average F1 of 64.9. The gap in TPR@1\%FPR is even more striking: DIRE's TPR falls to near zero across nearly all OOD generators and sequence lengths, making it effectively useless for high-confidence detection. As discussed in \cref{subsec:crossgenerator}, this failure occurs because DIRE's detection cue measures distance to the reference generator's distribution. Once the test samples come from an unseen generator, the reconstruction errors of synthetic and real samples become indistinguishable, as shown by the overlapping distributions in \Cref{subfig:sssd_dire_mse_ettm_64,subfig:sssd_dire_mse_weather_128}. Disjoint-CNN, in contrast, learns synthetic artifacts directly from the raw signal and maintains stable performance across unseen generators. These results demonstrate the viability of black-box detection for diffusion-generated time series, and motivate further exploration of black-box time series detectors.


\textbf{Impact of time series length.} Performance trends with sequence length differ between ID and OOD evaluation. As shown in \cref{tab:main_exp}, Disjoint-CNN retains near-perfect ID performance across all sequence lengths, including 94.7 F1 at 128-length on SSSD samples. However, in OOD, its average F1 drops from 84.8 at 32-length to 69.1 at 128-length. This pattern indicates that the limitation at longer sequences is \textit{generalization} rather than detectability: generator-specific synthetic artifacts remain easy to identify when the detector is tested on the same generator it was trained on, but the artifacts that transfer across generators become harder to capture as the sequence grows. The same pattern holds across the other black-box classifiers we compared (InceptionTime and LITE; see \cref{tab:exp_sssd} in \cref{appx:classifiers}), all of which show strong ID performance at 128-length but a marked OOD drop. DIRE declines more gradually with length but starts from a much lower OOD baseline, and remains the weakest detector at every sequence length.

\textbf{Impact of synthetic data quality.} As shown in \cref{fig:quality_detectability}, both detectors struggle more as generation quality improves. Higher synthetic data quality leaves fewer artifacts for the detector to identify, therefore making detection harder. Diffusion-TS consistently produces the highest-quality synthetic data across datasets, and correspondingly, both DIRE and Disjoint-CNN achieve their weakest OOD performance against it. For instance, at 32-length sequences, Disjoint-CNN's F1 drops to 71.7 on Diffusion-TS compared to 98.5 on WaveStitch. Even so, Disjoint-CNN degrades more gracefully than DIRE, as it retains a higher TPR and AUC on Diffusion-TS, while DIRE's TPR@1\%FPR drops to zero. This further shows that learning from raw signals is more reliable than reconstruction-based detection when the generation quality is high.
\section{Conclusion}

We present the first systematic exploration of white-box and black-box detection of diffusion-generated time series. The white-box approach, a reconstruction-based detector adapted from the image domain, works well in the in-distribution but breaks down under generator shift because the time series domain lacks a generic generator that could serve as a universal reconstruction prior. In contrast, an off-the-shelf classifier used as a black-box detector achieves an average OOD F1 of 79.2, a 22.1\% relative improvement over the white-box approach, and a TPR@1\%FPR of 57.2. Diffusion-generated time series detection is therefore not a direct transfer of the image domain problem, and black-box detection is a viable starting point.

\subsection{Recommendations for Future Work}

Black-box cross-generator detection remains poorly understood. We provide a first working framework, and several open questions follow from our results:

\begin{enumerate}[leftmargin=2em]
    \item \textbf{What signals do black-box classifiers capture?} Identifying which synthesis artifacts the classifier relies on could inform new detector designs that explicitly leverage these signals, and may reveal whether different diffusion models leave structurally similar fingerprints.
    \item \textbf{Can a black-box detector handle both autoregressive and diffusion generators?} Our study focuses on diffusion-based generators. Extending coverage to include autoregressive models would broaden practical utility, especially as time series large models proliferate.
    \item \textbf{How does sequence length shape generalization?} At 128-length, our detector retains strong in-distribution performance but loses much of its OOD advantage, suggesting that the artifacts which transfer across generators become harder to capture as sequences grow.
\end{enumerate}

Beyond cross-generator detection, the broader problem of cross-dataset detection remains unexplored. Generalizing from one dataset to another, rather than across generators within the same dataset, is of vital importance to practitioners and a natural next step.

\section*{GenAI Usage Disclosure}
Generative AI models were used solely for language polishing and grammar correction during paper writing.







\bibliographystyle{ACM-Reference-Format}
\bibliography{refs}

\clearpage
\crefalias{section}{appendix}
\crefalias{subsection}{appendix}
\appendix

\section{Experimental Settings} \label{appx:experimental_settings}

\textbf{Diffusion-based time series generators.} We describe the four diffusion-based time series generators considered in our experiments. \textit{SSSD}~\citep{DBLP:journals/tmlr/AlcarazS23} is a diffusion model that uses S4~\citep{DBLP:conf/iclr/KongPHZC21} layers as the denoising backbone. Originally proposed for time series imputation and forecasting, its iterative denoising procedure also supports unconditional sample generation, and it operates directly in data space rather than through a latent representation. \textit{TSDiff}~\citep{DBLP:conf/nips/KolloviehABZ0023} is a self-guiding diffusion model for probabilistic time series forecasting that avoids the need for an auxiliary classifier by leveraging observation self-guidance during sampling. The same backbone supports three modes, such as predict, refine, and synthesize, and we use it as an unconditional generator in this work. \textit{Diffusion-TS}~\citep{DBLP:conf/iclr/YuanQ24} is an interpretable diffusion model for time series generation built on a transformer backbone, which decomposes time series into trend and seasonal components and uses a Fourier-based loss to capture temporal semantics, often achieving the highest sample quality among the generators we consider. Finally, \textit{WaveStitch}~\citep{DBLP:journals/pacmmod/ShankarCDH25} is a conditional diffusion model designed for fast and flexible time series generation that produces long sequences by stitching together overlapping chunks generated via parallel denoising, and it supports a variety of conditioning signals.

\textbf{Evaluation metrics.} We evaluate detectors using five standard binary classification metrics, for all of which higher values indicate better performance. The \textit{F1} score is the harmonic mean of precision and recall, balancing correct classifications against misclassifications across both classes. \textit{Accuracy} is the fraction of samples correctly classified as real or synthetic. \textit{Average Precision} (AP) is the area under the precision-recall curve, summarizing detection performance across all decision thresholds. \textit{AUC} is the area under the receiver-operating-characteristic curve, measuring the detector's ability to rank and separate real from synthetic samples across all thresholds. Finally, \textit{TPR@1\%FPR}, the True Positive Rate at 1\% False Positive Rate, measures detector sensitivity at a strict low false-positive operating point. We report this metric because practical detectors typically operate under tight false-positive constraints, and performance at a fixed FPR captures that regime directly.

\textbf{Quality metrics.} We use four metrics from prior work to evaluate synthetic time series quality, where lower values indicate higher quality. \textit{Context-FID}~\citep{DBLP:conf/iclr/JehaBMKNFGJ22} adapts the Fr\'{e}chet Inception Distance to time series, measuring distributional similarity between real and synthetic samples in a learned feature space. The \textit{correlational}~\citep{DBLP:journals/corr/abs-2006-05421} score is the absolute difference between their cross-channel correlation matrices, capturing inter-channel dependencies. The \textit{discriminative}~\citep{DBLP:conf/nips/YoonJS19} score reports the deviation from chance (0.5) of a binary classifier trained to separate real from synthetic samples, with values near 0 indicating indistinguishability. Finally, the \textit{predictive}~\citep{DBLP:conf/nips/YoonJS19} score trains a forecaster on synthetic data and evaluates it on real data, with low values indicating preserved temporal structure.

\section{Black-box Architecture Comparison} \label{appx:classifiers}
\cref{tab:exp_sssd} compares Disjoint-CNN~\citep{DBLP:conf/incdm/FoumaniTS21} against two other strong time series classifiers, InceptionTime~\citep{DBLP:journals/datamine/FawazLFPSWWIMP20} and LITE~\citep{DBLP:conf/dsaa/IsmailFawazDBWF23}, across all four generators and three sequence lengths.

\begin{table}
\centering
\caption{All metrics are higher the better, with TPR measured at a 1\% FPR. Values in \textcolor{gray}{gray} are ID, while values in black are OOD. The best average OOD scores are highlighted in \colorbox{green!15}{green}.}
\label{tab:exp_sssd}
\resizebox{\linewidth}{!}{%
\begin{tabular}{llccccc}

\toprule

Generator & Method & F1 $\uparrow$ & Acc. $\uparrow$ & AP $\uparrow$ & AUC $\uparrow$ & TPR $\uparrow$ \\

\midrule
\multicolumn{7}{c}{\cellcolor{blue!15}{\textit{32-length}}} \\
\midrule

\multirow{3}{*}{SSSD} & InceptionTime & \textcolor{gray}{98.9} & \textcolor{gray}{99.0} & \textcolor{gray}{99.9} & \textcolor{gray}{99.9} & \textcolor{gray}{98.4} \\
& LITE & \textcolor{gray}{96.9} & \textcolor{gray}{97.6} & \textcolor{gray}{99.9} & \textcolor{gray}{99.9} & \textcolor{gray}{98.4} \\
& Disjoint-CNN & \textcolor{gray}{99.7} & \textcolor{gray}{99.7} & \textcolor{gray}{100.} & \textcolor{gray}{100.} & \textcolor{gray}{99.8} \\
\cmidrule{1-7}
\multirow{3}{*}{TSDiff} & InceptionTime & 78.5 & 69.1 & 77.9 & 70.2 & 56.5 \\
& LITE & 80.4 & 73.4 & 89.1 & 87.1 & 59.9 \\
& Disjoint-CNN & 84.1 & 77.8 & 95.2 & 95.4 & 76.5 \\
\cmidrule{1-7}
\multirow{3}{*}{Diffusion-TS} & InceptionTime & 69.8 & 57.0 & 75.2 & 69.5 & 35.9 \\
& LITE & 68.2 & 55.4 & 75.0 & 70.7 & 31.2 \\
& Disjoint-CNN & 71.7 & 59.2 & 61.9 & 57.8 & 19.6 \\
\cmidrule{1-7}
\multirow{3}{*}{WaveStitch} & InceptionTime & 98.1 & 98.1 & 99.7 & 99.6 & 97.4 \\
& LITE & 96.5 & 97.1 & 99.6 & 99.5 & 97.9 \\
& Disjoint-CNN & 98.5 & 98.3 & 99.9 & 99.9 & 99.2 \\
\cmidrule{1-7}\morecmidrules\cmidrule{1-7}
\multirow{3}{*}{Avg. OOD} & InceptionTime & 82.2 & 74.8 & 84.2 & 79.8 & 63.3 \\
& LITE & 81.7 & 75.3 & \cellcolor{green!15}{87.9} & \cellcolor{green!15}{85.8} & 63.0 \\
& Disjoint-CNN & \cellcolor{green!15}{84.8} & \cellcolor{green!15}{78.4} & 85.7 & 84.4 & \cellcolor{green!15}{65.1} \\

\midrule
\multicolumn{7}{c}{\cellcolor{blue!15}{\textit{64-length}}} \\
\midrule

\multirow{3}{*}{SSSD} & InceptionTime & \textcolor{gray}{98.5} & \textcolor{gray}{98.6} & \textcolor{gray}{100.} & \textcolor{gray}{100.} & \textcolor{gray}{100.} \\
& LITE & \textcolor{gray}{99.5} & \textcolor{gray}{99.6} & \textcolor{gray}{100.} & \textcolor{gray}{100.} & \textcolor{gray}{99.9} \\
& Disjoint-CNN & \textcolor{gray}{99.9} & \textcolor{gray}{99.9} & \textcolor{gray}{99.9} & \textcolor{gray}{100.} & \textcolor{gray}{98.6} \\
\cmidrule{1-7}
\multirow{3}{*}{TSDiff} & InceptionTime & 80.8 & 73.2 & 72.5 & 61.5 & 56.6 \\
& LITE & 84.4 & 77.5 & 87.5 & 82.8 & 78.8 \\
& Disjoint-CNN & 84.6 & 77.7 & 98.9 & 99.2 & 86.4 \\
\cmidrule{1-7}
\multirow{3}{*}{Diffusion-TS} & InceptionTime & 68.5 & 55.0 & 75.4 & 68.7 & 35.4 \\
& LITE & 68.6 & 54.2 & 74.5 & 69.1 & 32.9 \\
& Disjoint-CNN & 71.5 & 59.3 & 67.6 & 65.5 & 15.7 \\
\cmidrule{1-7}
\multirow{3}{*}{WaveStitch} & InceptionTime & 92.7 & 90.7 & 93.1 & 89.9 & 86.7 \\
& LITE & 92.1 & 88.9 & 99.4 & 99.5 & 89.9 \\
& Disjoint-CNN & 95.4 & 93.7 & 97.2 & 98.1 & 81.6 \\
\cmidrule{1-7}\morecmidrules\cmidrule{1-7}
\multirow{3}{*}{Avg. OOD} & InceptionTime & 80.6 & 73.0 & 80.3 & 73.4 & 59.6 \\
& LITE & 81.7 & 73.5 & 87.1 & 83.8 & \cellcolor{green!15}{67.2} \\
& Disjoint-CNN & \cellcolor{green!15}{83.8} & \cellcolor{green!15}{76.9} & \cellcolor{green!15}{87.9} & \cellcolor{green!15}{87.6} & 61.2 \\

\midrule
\multicolumn{7}{c}{\cellcolor{blue!15}{\textit{128-length}}} \\
\midrule

\multirow{3}{*}{SSSD} & InceptionTime & \textcolor{gray}{99.5} & \textcolor{gray}{99.5} & \textcolor{gray}{100.} & \textcolor{gray}{100.} & \textcolor{gray}{100.} \\
& LITE & \textcolor{gray}{99.8} & \textcolor{gray}{99.8} & \textcolor{gray}{100.} & \textcolor{gray}{100.} & \textcolor{gray}{100.} \\
& Disjoint-CNN & \textcolor{gray}{94.7} & \textcolor{gray}{96.5} & \textcolor{gray}{100.} & \textcolor{gray}{100.} & \textcolor{gray}{100.} \\
\cmidrule{1-7}
\multirow{3}{*}{TSDiff} & InceptionTime & 66.8 & 50.6 & 69.1 & 56.5 & 42.5 \\
& LITE & 67.0 & 50.8 & 66.7 & 53.3 & 42.9 \\
& Disjoint-CNN & 63.7 & 49.8 & 81.9 & 80.0 & 52.7 \\
\cmidrule{1-7}
\multirow{3}{*}{Diffusion-TS} & InceptionTime & 66.4 & 49.8 & 58.5 & 45.7 & 21.7 \\
& LITE & 66.6 & 50.1 & 50.4 & 36.7 & 10.7 \\
& Disjoint-CNN & 62.8 & 46.9 & 46.7 & 36.2 & 10.2 \\
\cmidrule{1-7}
\multirow{3}{*}{WaveStitch} & InceptionTime & 79.3 & 70.2 & 92.1 & 90.1 & 82.7 \\
& LITE & 75.6 & 64.1 & 86.5 & 81.4 & 77.3 \\
& Disjoint-CNN & 80.8 & 75.9 & 93.5 & 93.0 & 73.1 \\
\cmidrule{1-7}\morecmidrules\cmidrule{1-7}
\multirow{3}{*}{Avg. OOD} & InceptionTime & \cellcolor{green!15}{70.9} & 56.9 & 73.2 & 64.1 & \cellcolor{green!15}{48.9} \\
& LITE & 69.7 & 55.0 & 67.9 & 57.1 & 43.6 \\
& Disjoint-CNN & 69.1 & \cellcolor{green!15}{57.5} & \cellcolor{green!15}{74.1} & \cellcolor{green!15}{69.7} & 45.3 \\

\bottomrule
\end{tabular}
}
\end{table}

\end{document}